\documentclass[a4paper, fleqn]{cas-dc}

\usepackage{makecell}
\usepackage{svg}
\usepackage[authoryear]{natbib}
\usepackage{amsmath}
\usepackage{multicol}
\usepackage{multirow}
\usepackage{longtable}

 \usepackage{pdflscape}
\def\tsc#1{\csdef{#1}{\textsc{\lowercase{#1}}\xspace}}
\tsc{WGM}
\tsc{QE}

\begin{document}
\let\WriteBookmarks\relax
\def\floatpagepagefraction{1}
\def\textpagefraction{.001}

\shorttitle{Entity Identifier: A Natural Text Parsing-based Framework For Entity Relation Extraction}    


\title [mode = title]{Entity Identifier: A Natural Text Parsing-based Framework For Entity Relation Extraction} 

\tnotemark[1] 

\tnotetext[1]{<tnote text>} 


\author[1]{El Mehdi Chouham}[]

\ead{elchouham@novelis.io}


\credit{}

\author[1]{{Jessica} {López Espejel}}[
    orcid=0000-0001-6285-0770
]



\ead{jlopezespejel@novelis.io}



\affiliation[1]{organization={Novelis
Research and Innovation Lab},
            addressline={207 Rue de Bercy}, 
            city={Paris},
            postcode={75012}, 
            state={},
            country={France}}

%

\author[1]{{Mahaman Sanoussi} {Yahaya Alassan}}[]


\ead{syahaya@novelis.io}


\credit{}

\author[1]{Walid Dahhane}[orcid=0000-0001-5387-3380]

\ead{wdahhane@novelis.io}


\credit{}


\author[1]{El Hassane Ettifouri}[
    orcid=0000-0001-5299-9053
]

\ead{eettifouri@novelis.io}


\cormark[1]
\credit{}


\cortext[1]{Corresponding author}

\cortext[1]{Corresponding author}

\fntext[1]{}
\begin{abstract}
 The field of programming has a diversity of paradigms that are used according to the working framework. While current neural code generation methods are able to learn and generate code directly from text, we believe that this approach is not optimal for certain code tasks, particularly the generation of classes in an object-oriented project. Specifically, we use natural language processing techniques to extract structured information from requirements descriptions, in order to automate the generation of CRUD (Create, Read, Update, Delete) class code. To facilitate this process, we introduce a pipeline for extracting entity and relation information, as well as a representation called an "Entity Tree" to model this information. We also create a dataset to evaluate the effectiveness of our approach.
\end{abstract}



\begin{highlights}

   \item We have presented Entity Identifier, a pipeline method for transforming requirements specifications in natural language into a model diagram that incorporates Stanford Scene Graph Parsing.

   \item We create a dataset and define evaluation metrics to assess the effectiveness of our approach and facilitate future research in this area.

    \item Our method achieves high scores on simple requirement statements, but struggles in handling complex Wikipedia paragraphs.

\end{highlights}


\ExplSyntaxOn
\keys_set:nn { stm / mktitle } { nologo }
\ExplSyntaxOff

\maketitle

\section{Introduction}
\label{sec: introduction}

    In Natural Language Processing (NLP), many tasks put effort on converting input text into a more readily understandable form for humans. Examples of such tasks include translation \citep{vaswani2017_attention}, summarization \citep{zhang2019_pegasus, espejel2021_saucissonnage}, question answering \citep{codeT5}, text rephrasing \citep{laugier2021_rephrases}, and named entity recognition \citep{tan2023_named_entity}. However, only a relatively small number of tasks, such as sequence and token classification, may be primarily useful for machines. We believe that the development of automatic methods to model natural language for machine use has the potential to enable a wide range of future applications. These models may allow for the creation of systems that can understand and interpret human language and leverage this knowledge for downstream use, potentially leading to new and innovative applications in various industries and sectors.

    For instance, in the field of text-to-code generation, current AI models such as CodeBERT \citep{codebert}, CodeT5 \citep{codeT5}, JaCoText \citep{JaCoText}, and Codex \citep{codex} have shown promising results, but still struggle with  lack of optimization, inconsistency, and syntax issues. The latest is a major limitation, as syntactically correct code is necessary for a machine to execute it. Additionally, practical considerations such as code structure, clearness, and readability are important aspects of software development that these models have not yet been able to fully address. We believe that incorporating a structuring phase for natural language inputs could significantly advance the capabilities of tasks like text-to-code generation.

        In this work, we work on automating the job of requirement analysis by extracting key information that can be directly used to build UML (United Modeling Language) Class diagram and generate class code for CRUD (Create, Read, Update, Delete) operations from natural language requirement specifications. Our primary goal is exploring the benefits of structuring natural language for code generation. Specifically, we aim to extract entities and relationship triplets, including their characteristics, in a manner similar to the joint entity and relation extraction task \citep{relation_extraction}. In addition, we  aim to extract common unnamed entities, data types, class attributes, and database relations between entities, in order to build a rich representation, we refer to as an Entity Tree. This representation can be useful for downstream tasks. Figure 1 illustrates an example of the Entity Tree representation.

    In development workflows, diagrams such as UML and MCD (Merise Conceptual Diagram) are extensively used for engineering and visualization purposes. These diagrams enable collaborators to analyze, refine, and plan different aspects of a project more effectively. Our approach not only grants direct access to these diagram types but also simplifies the generation of class code and database structure using template heuristics. This liberates developers from repetitive tasks, allowing them to concentrate on more  challenging ones. The key advantage of this approach in code generation task is the creation of more dependable code with fewer syntax errors.

    In this work, we present a parsing framework that constructs an Entity Tree (ET) from natural language text. To the best of our knowledge, there are currently no datasets available for common entity and relation extraction. Therefore, we create a dataset and define evaluation metrics to assess the effectiveness of our approach and facilitate future research in this area. Our method achieves notable results in the overall extraction of triplets.


     The rest of the paper is organized as follows: In section \ref{sec:2_related_work}, we present the related work. In section \ref{sec:3_Problem_Formulation}, we provide a comprehensive formulation of the problem. In Section \ref{sec:4_Proposed_Method}, we describe in details our proposed method. In Section \ref{sec:5_experimental_protocol}, we present the experimental protocol and the proposed dataset. In Section \ref{sec:6_results}, we provide obtained results and discuss them. Finally, We conclude in  Section \ref{sec:7_conclusion} and present and some future directions.

    \begin{figure*}[h!]
        \centering
        \includegraphics[width=0.8\textwidth]{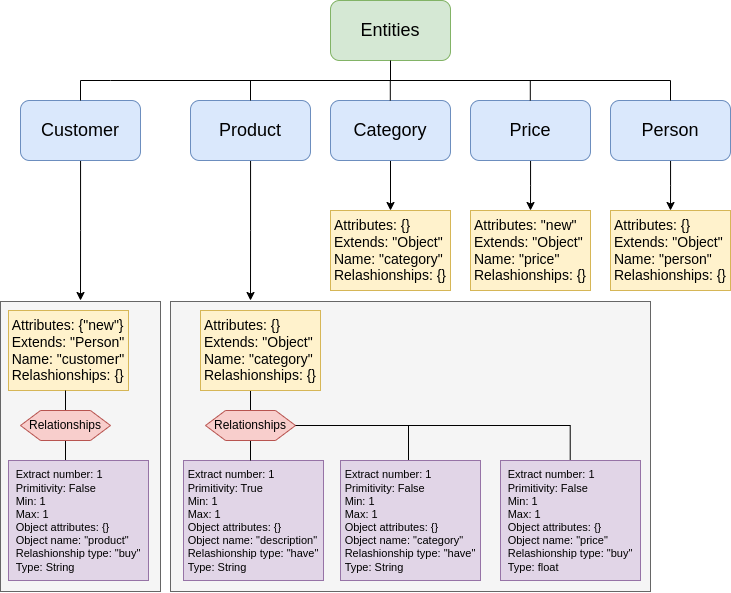}
        \caption{Entity Tree representation of the sentence : \textit{``A new customer buys a product. It has a text description, a category and a price. A customer is person."} with a focus on the relation (customer, buy, product)}
       \label{fig:Entity_Tree_example}
    \end{figure*}

\section{Related Work}
\label{sec:2_related_work}


    There is a significant amount of prior research on system requirements parsing, but most existing models do not anticipate the use of an entirely natural language input. \cite{patterns} proposed a framework for requirement parsing that utilizes a set of word processing tools and typing rules. This approach involves the user in typing the input, with the system providing error messages and highlighting patterns to enforce adherence to what the paper refers to as a \emph{controlled natural language for requirements' specifications}. This approach relies on the user to follow specific typing rules and may not be suitable for users who want a fully natural user experience while typing the input.

    In \cite{generatingUML}, a greater emphasis is placed on the use of natural language processing and ontology techniques with OpenNLP to build the parsing architecture. This approach allows for more flexibility for the user, but the parsers rely on transforming complex sentences into simple sequences rather than consistently representing the information. Additionally, the approach does not utilize coreference resolution, meaning that splitting sentences can result in loss of linking information. Furthermore, the use of rules to ignore certain concept words later in the pipeline may lead to error propagation and confusion. Overall, this approach may require more user intervention to ensure accurate parsing of the input.
\newline

    Similarly, \citet{AmrutaAmune2016ReflectingNL} uses a suite of natural language processing tools to extract UML diagrams directly. Their approach involves the use of heuristic rules for identifying classes, relationships, and objects. However, our research posits that it is more effective to extract entities and relationships jointly, as the semantics of a sentence are dependent on the combination of both. Furthermore, while the aforementioned tool is presented as software specifically designed for the creation of UML diagrams, our research aims to output the extracted elements in a more widely-used representation.

    The approach presented in \citet{extractingDomain} addresses the issue of information loss through the use of a coreference resolver in the pipeline. The pipeline utilizes both dependency parsing and syntactic parsing during the parsing phase, and employs a set of rules to construct a Domain Model in a subsequent phase. However, the approach has some limitations, including the fact that coreference resolution is performed after segmentation, and the reliance on noun phrases and verbs instead of a compact representation for building the domain model.

    In our research, we focus on a fundamental task of common entity and relation extraction. Unlike traditional entity and relation extraction, our task aims to extract all entities and relationships in a text, not just named entities. We use the extracted sets of entities and relationships to build a model of the system. We believe that this approach can highlight the potential utility of jointly extracting unnamed entities and relations as a distinct task within the field of knowledge extraction. After optimizing our pipeline for this task, we then build our domain model equivalent referred to as the Entity Tree.

    \section{Problem Formulation}
    \label{sec:3_Problem_Formulation}

    Our goal is to design a mechanism that translates a natural language description of system requirements into a structured representation called an Entity Tree (as shown in Figure \ref{fig:Entity_Tree_example}). This representation will enable us to easily generate UML, MCD, CRUD diagrams, and database structures. 

    Let $S$ be a sequence and $E=\{e_1, e_2, ..., e_{|S|}\}$ the set of its entities' components. An Entity Component $e_1$ is defined by 4 elements: 

    \begin{itemize}
       \item \textbf{Attributes - } set of descriptive labels representing the entity.
       \vspace{-0.1cm}
       
       \item \textbf{Extends -} the entity it extends, if it has a parent entity.
       \vspace{-0.5cm}
       
       \item \textbf{Name -} the entity name.
       \vspace{-0.1cm}
       
       \item \textbf{Relationships -} set of relationships with other entities.
    \end{itemize}

    
    We define a relationship as the link between two entities. Here we refer to the entity containing the list of relationships as the subject $e_s$, the target entity as the object $e_o$ and the predicate as the relation $r$.  We explicitly define a relationship as a dictionary of the following eight items:
    
    \begin{itemize}
       \item \textbf{Extract number -} the exact number of the subject entities $e_s$, if mentioned.
       \vspace{-0.1cm}
       
       \item \textbf{Primitivity -} it indicates if the $e_o$ is primitive. An entity is primitive if it is a literal such as a number, a date, a boolean, etc. In other words, when it is not a \textit{Class} instance. Values which are considered primitive can be redefined by the user.
       \vspace{-0.1cm}
       
       \item \textbf{Min -} cardinality on the $e_s$ side.
       \vspace{-0.1cm}
       
       \item \textbf{Max -} cardinality on the $e_o$ side.
       \vspace{-0.1cm}
       
       \item \textbf{Object attributes -} set of labels describing $e_o$.
       \vspace{-0.1cm}
       
       \item \textbf{Object name -} the $e_o$ name.
       \vspace{-0.1cm}
       
       \item \textbf{Relationship type -} the predicate  $r$.
       \vspace{-0.1cm}
       
       \item \textbf{Type -} the type notion refers to the parent class of $e_o$ if it has one, otherwise it assigns a corresponding primitive type.
       
    \end{itemize}
    
    The Entity Tree models the entities and relationships of a sentence in a very rich way as it describes every item and its relation with other items. Its data structure serves as a representation of entities and their relationships within a sentence or piece of text. Its utilization allows for a more intricate and comprehensive depiction of sentence structure, surpassing conventional approaches such as parsing trees. In the example \textit{"The cat sat on the mat."}, the words \textit{"cat"} and \textit{"mat"} are entities, and the relationship between them is \textit{"sat on"}.   
    In our case, the Entity Tree can also include additional information about each entity, such as its type or attributes. This presentation offers valuable applications in tasks such as information extraction, text summarization, and question answering. 
    

    \section{Proposed method}
     \label{sec:4_Proposed_Method}


    
\begin{figure}[h!]
    \centering
      \includegraphics[width=0.5\textwidth, height=10cm]{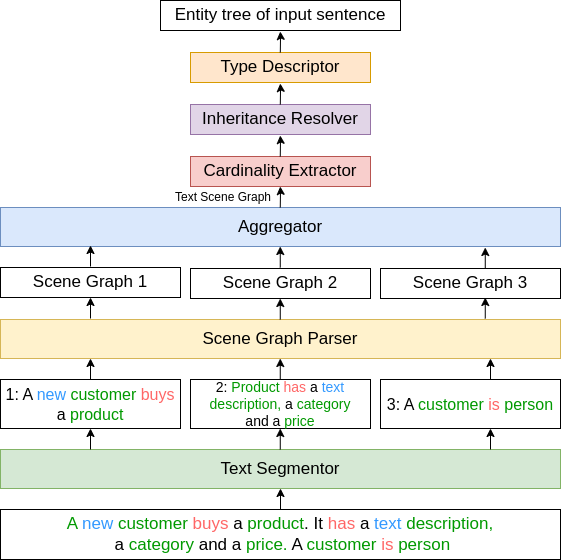}
      \caption{Pipeline architecture. \textit{Best viewed in color.}}
      \label{fig:pipeline_architecture}
    \end{figure}

    Our approach uses a combination of multiple natural language processing (NLP) tools to extract entities, relationships, and their characteristics. An overview of our pipeline is shown in Figure \ref{fig:pipeline_architecture}.    Firstly, the input text undergoes coreference resolution to replace each pronoun with the word to which it refers. Secondly, the text is segmented into sentences. Thirdly, for each sentence, we use a Scene Graph Parser (SGP)\footnote{A Stanford's CoreNLP tool}  \citep{sgp} to obtain the corresponding scene graph. An aggregator is used to fuse different scene graphs. Finally, we apply additional methods to double-check features extraction, generate cardinality, and build up the Entity Tree representation of the input. We detail each component of our system in the following subsections.
    

    \subsection{Text Segmentation}
    To use the CoreNLP scene graph parser for building a scene graph, the input should adhere to the scene graph parser’s input format. According to the official documentation, the implementation only works on single sentences. For this reason, the input goes through sequential methods, a sentence segmentation function and a coreference resolver.
    
    Firstly, the text goes through Spacy’s integrated sentence sequencer. The latter ensures that the text will not only be split based on dots. Instead, the split is done using regular expressions while taking into consideration links, phone numbers, decimal numbers, etc. 
    
    Secondly, pronouns are replaced by the nominal group to which they refer to in sentences starting with a pronoun. This prevents sending to the scene graph parser sentences with no nominal groups and spread confusing downstream. 
    
    The text segmentation block outputs a list of clear sentences, that can be semantically understood without the need for the rest of the text. In the next step, each sentence will be sent to the scene graph parser.
    
    \subsection{Scene Graph Parsing}
    
    The scene graph parser extracts knowledge from a sentence into a scene graph. As its name suggests, the scene graph can be thought of as a representation of the scene described in the input sentence. Various implementations are available. In our work, we stick with the rule-based scene graph parser since its implementation showed the best stability. Practically, the CoreNLP implementation returns a list of relationships. A relationship consists of a subject, the predicate and the object, and a list of attributes built similarly with a list of sets, each containing a subject, the predicate and an attribute.
    
    Scene graph parser starts by building an enhanced version of a dependency tree of the sentence (Klein and Manning, 2003), only then it resolves quantificational modifiers, coreferences and plural.
    
    \subsubsection{Quantificational modifiers}
    
    Dependency tree parsers treat noun expression quantificational modifiers such as \textit{“both of”} or \textit{“a bunch of”} like noun phrases leading to the latent dependency trees containing root pronouns instead of the subject. Scene graph parser tackles this issue by checking if a noun expression is a quantification modifier against a precompiled list. This addition guarantees an intermediate dependency tree were the head component is an entity.
    
    \subsubsection{Coreference}
    
    The parser performs a coreference resolution on the pronouns of the dependency tree, disambiguating pronouns and other words that might be ambiguous without context. It uses an adaptation of the \citet{hobbs} algorithm for dependency trees, to save underlying semantic link between sentences, clear up any confusions and improve the accuracy of downstream natural language processing tasks.
    
    \subsubsection{Plural resolution}
    
    In the context of plural resolution, the scene graph parser is tasked with addressing the challenge of "collective-distributive" ambiguity, which refers to the difficulty of determining whether a group noun refers to a single entity that is comprised of multiple individual parts (collective) or multiple distinct entities (distributive). On the one hand, in the example sentence \textit{"Two guys in police uniforms"}, the parser would need to determine that the noun \textit{"guys"} refers to two distinct individuals, each of whom is wearing a police uniform. On the other hand, in the example sentence \textit{"Two passengers in the backseat of the car"}, the parser would need to recognize that the noun "passengers" refers to two distinct individuals, but that there is only one backseat entity. The scene graph parser only considers that the distributive plural is far more common. This is convenient as it aligns with the common nature of specifications' descriptions.
    
    At this stage the parser outputs what is referred to as a Semantic Graph. From this point, the semantic graph will undergo a set of methods to extract objects and attributes.
    
    \subsubsection{Rule-based Parser}
    
    The goal of these procedure is to extract object entities, relations, and attributes from the input sentence using a semantic graph and Semgrex2 expressions. As of the time this research was conducted, Semgrex2 expressions were able to capture a wide range of pattern constructions. It supports the following patterns: 
    \begin{itemize}
    \setlength\itemsep{0.01cm}
        \item Adjectival modifiers
        \item Subject-predicate-object constructions 
        \item Subject-predicate constructions 
        \item Copular constructions
        \item Prepositional phrases
        \item Possessive constructions
        \item Passive constructions
        \item Clausal modifiers of nouns
    \end{itemize}
    
    In addition to the rule-based approach, the CoreNLP library offers a classifier-based approach for extracting relations between entities. However, we have found that the rule-based approach is more reliable, as the classifier-based approach may overlook complex patterns and underlying semantics. The output of this process is a scene graph, which consists of a list of relationships and a list of attributes. 
    
    We transform the scene graph into a list of entities, each of which has a sub-list of attributes and a sub-list of relationships. A relationship is a link between two entities, consisting of an object entity and a predicate verb describing the connection between the subject entity and the object entity.
    
    \subsection{Cardinality Extraction}
    
    The scene graph produced at this stage does not include information about the cardinality of entities, as the process converts the corresponding words into their lowercase stem. To address this issue, our cardinality extractor iterates through the list of relationships and the corresponding sentences to determine cardinality. Plural noun expressions are assigned a cardinality of "*", while singular noun expressions are assigned a cardinality of "1". Additionally, for database structure purposes, if an object is described as "optional" in the sentence, the corresponding entity is assigned a cardinality of "0".\\
    
    This will help easily deduce database cardinality types' downstream if needed. For example, in \textit{"A level has multiple bosses"} we can deduce a many-to-one type of cardinality. Furthermore, this module also extracts and preserves any indicated numerical quantifiers as  the "Exact Number" entry of the Entity Tree.

    \subsection{Inheritance Resolver}
    
    Attributes in a scene graph can be either noun expressions or adjectives. However, in our case, for a noun attribute to be considered as an adjective, it should not be mentioned else where as an entity, otherwise it will primarily be considered as a parent class to the described entity instead of an attribute.
    
    We implement an engine that uses a pre-defined set of rules to iterate through scene graph attributes. It looks up for relationships containing a \textit{“to be”} predicate and applies rules to determine the entity's relative position in the inheritance hierarchy. In this way, we to build up the inheritance.
    
    \subsection{Type description}
    Similar to a variable type in programming, we assign to entities the kind of data that they can hold. 
    If there are types explicitly described through specific attributes appearing in a list pre-defined by the user (such as \textit{Date, Long, Short,} and {Int}), the corresponding entity is marked as primitive and is assigned this type. Object Entities which appear as subjects elsewhere in the text are not assigned a type. Thus, their value is defaulted to \textit{"String"}.
    
\begin{figure}[h!]
      \includegraphics[width=\linewidth]{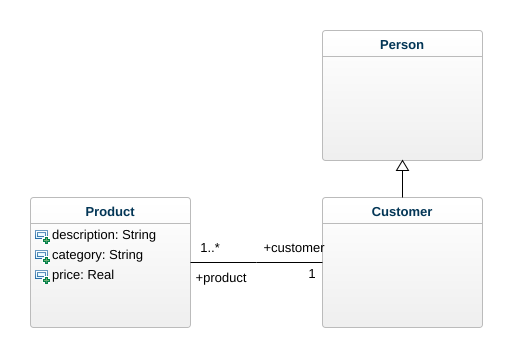}
      \caption{UML class diagram of the sentence: ``A new customer buys a product``. It has a text description, a category and a price. A customer is person.}
      \label{fig:class-diagram}
    \end{figure}

    The Entity Tree produced reflects domain model-like information, including class attributes and relationships with additional entries. Heuristics can be applied to this Entity Tree to generate consistent class code or UML class diagrams from directly system requirements. With simple rules script, we easily generate the UML class diagram illustrated in Figure \ref{fig:class-diagram}.
    
\section{Experimental Protocol}
 \label{sec:5_experimental_protocol}

 In this section, we will present our dataset, as well as the metrics used to evaluate our method. 

\subsection{Dataset}
    \subsubsection{Entity relation triplet dataset}
    We aim to evaluate our entity relation extraction work on a dataset specifically tailored for this purpose. However, we have found that existing datasets such as WebNLG \citep{webnlg} and New York Times \citep{Nyt} primarily concentrate on the extraction of named entities, which is not appropriate for our use case.  For this reason, we constructed a dataset consisting of randomly selected Wikipedia paragraphs and labeled each paragraph with its corresponding $(e_s, relationship, e_o)$ triple for evaluation purposes. Our dataset, contains :
    \begin{itemize}
    \setlength\itemsep{0.01cm}
        \item Text theme
        \item The text content
        \item The corresponding $(e_s, r, e_o)$ triplets
        \item Scene Graph Extracted Dependencies
    \end{itemize}
    The dataset contains a total of 198 examples. The dataset is only intended for evaluation. We therefore consider that this amount of examples is sufficient to evaluate an algorithm that does not contain a training phase.
    
    \subsubsection{Patterns set}
    A close examination of Wikipedia paragraphs reveals that they can be quite complex, which is at odds with the more precise, concise, and syntactically clear nature of system requirements descriptions. To evaluate the performance of our parser on a more representative way, we have developed an additional checklist of six patterns.

\subsection{Metrics}
For evaluation, we define some metrics to evaluate the extraction accuracy. We believe that a successful information extraction of the sentence resides in a consistent extraction of the triplet $(e_s, r, e_o)$. Therefore, we focus on evaluating this aspect. We propose metrics based on Intersection over Union as well as BLEU \citep{bleu} score. The approach of using BLEU is motivated by the idea that our task can be viewed as a type of translation, and the BLEU metric is well-suited to evaluating the quality of translations.

\paragraph{\textbf{\textbullet{~~Pair IoU -}}}
        First of all, we introduce an IoU metric to evaluate the number of exactly detected pairs.
        \begin{equation*}
            IoU_{Pair}(A, B) = \frac{A \cap B}{A \cup B}
        \end{equation*}
        where $A$ is reference entity pairs' set, and $B$ is candidate entity pairs' set. For the sake of readability, $e_s, e_o$ refer to the concatenation of the name and the attribute of each of the subject entity and the object entity respectively.

\paragraph{\textbf{\textbullet{~~Pairs mean BLEU -}}}
        We acknowledge that detecting exact pairs in complex text can be challenging, particularly when the entities are compound. Despite this, we believe that the use of the BLEU metric may provide a more suitable approach for comparing texts in this context compared to other methods. In this context, we introduce a BLEU mean of pairs metric.
        \begin{equation*}
            \resizebox{.5 \textwidth}{!} 
            {$
                BLEU_{Mean}(D_x, Dy) = \frac{
                \sum_{i \in D_x} \sum_{i \in D_y}\biggl[B(e_s^i, e_s^j) \geq k \biggl] \times \biggl[B(e_o^i, e_o^j) \geq k \biggl]
                }{	\biggl| D_x	\biggl| }
            $}
        \end{equation*}
        with :
        \begin{equation*}
            \resizebox{.5 \textwidth}{!}{$
            B(entity^i, entity^j) = \frac{BLEU(entity^i, entity^j) + BLEU(entity^j, entity^i)}{2}
            $}
        \end{equation*}
        
        $D_x$ is the reference sentences' set,
        $D_y$ is the candidate  sentences' set, and 
        $k$ is the threshold parameter. Comparisons with B() values that are less than k are omitted.

\paragraph{\textbf{\textbullet{~~Pairs exclusive BLEU -}}}

        We introduce another BLEU-based metric that only takes into account max BLEU value of the two permutations of  reference and generated pair.
        \begin{equation*}
            \resizebox{.5 \textwidth}{!} 
            {
               $BLEU_{Exclusive}(D_x, Dy) = \frac{
                \sum_{i \in D_x} \sum_{i \in D_y}\biggl[B(e_s^i, e_s^j) \geq k \biggl] \times \biggl[B(e_o^i, e_o^j) \geq k \biggl]
                }{	\biggl| D_x	\biggl| }
            $}
        \end{equation*}
        with :
        \begin{equation*}
            \resizebox{.5 \textwidth}{!}{$ 
            B(entity^i, entity^j) = max \biggl[BLEU(entity^i,\\ entity^j), BLEU(entity^j, entity^i)\biggl]
            $}
        \end{equation*}
        
       $entity$ refers to the concatenation of the name and the attribute of entity.

\paragraph{\textbf{\textbullet{~~Triplet BLEU -}}}
        Finally, we introduce a similar metric that evaluates both entities and their relation based on BLEU value of their concatenation.
        \begin{equation*}
           \resizebox{.5 \textwidth}{!} 
            {
                $BLEU_{Triplet}(D_x, Dy) = \frac{
                \sum_{i \in D_x} \sum_{i \in D_y}\biggl[B(W(triplet^i), W(triplet^j)) \geq k \biggl]
                }{	\biggl| D_x	\biggl| }$
            }
        \end{equation*}
        with :
        \begin{equation*}
           \resizebox{.5 \textwidth}{!}{$
            B(triplet^i, triplet^j) = max \biggl[BLEU(triplet^i, triplet^j), BLEU(triplet^j, triplet^i)\biggl]
            $}
        \end{equation*}
        where: 
        $W(.)$ is the concatenation of $e_s$, $r$ and $e_o$ of a triplet through white space characters, 
        $triplet$  refers to the concatenation of the name and the attribute of each entity and the relationship.      
        

\section{Results and discussion}
 \label{sec:6_results}

Our entity identification pipeline achieves high scores on the basic patterns set, with scores of 0.905, 0.929, and 0.929 using the metrics $IoU_{Pair}$, $BLEU_{Exclusive}@0.6$ and $BLEU_{Mean}@0.6$, respectively. It also performed exceptionally well in identifying triplets, earning a perfect score. These results demonstrate that the pipeline is capable of handling simple requirement statements, which is necessary for achieving our final goal.


However, when applied to complex Wikipedia paragraphs, the model encounters challenges and struggles to maintain the same level of performance. While the pipeline still manages to achieve decent results in terms of the overall extraction metric $BLEU_{Triplet}$ on the WikiTriplet dataset, it performs poorly when it comes to detecting entity pairs. Specifically, it obtains scores of 0.004, 0.036, and 0.036 on the $IoU_{Pair}$, $BLEU_{Exclusive}@0.6$, and $BLEU_{Mean}@0.6$ metrics, respectively, indicating a significant drop in performance for entity pair identification in complex text.


Our objective is to build a parser that performs well on simple sentences while also being adaptable to complex entries. For this reason, we have adopted an iterative optimization approach. This approach focuses on continuously improving the parser's performance based on its performance on the basic patterns list. The idea behind this strategy is to establish a solid basis in optimizing the pipeline's performance on simpler inputs and then gradually enhancing it to handle more complex text.

\begin{table}[hb!]
    \centering
    \begin{tabular}{|l|cc|}
         \hline
            & WikiTriplet dataset & Patterns' set \\
         \hline\hline
         $IoU_{Pair}$   & 0.004 & 0.905 \\
         \hline
         $BLEU_{Exclusive}@0.6$   & 0.036 & 0.929 \\
         \hline
         $BLEU_{Exclusive}@0.6$  & 0.036 & 0.929 \\
         \hline
         $BLEU_{Triplet}$    & 0.806 & 1.00 \\
         \hline
        \end{tabular}
    \caption{Evaluation of our parser on both datasets. $BLEU_{Exclusive}@0.6$ and $BLEU_{Exclusive}@0.6$ refer to the corresponding scores at a threshold of k=0.6}
    \label{tab:my_label}
\end{table}

\vspace{-1em}

\section{Conclusions and Perspectives}
 \label{sec:7_conclusion}
We have presented Entity Identifier, a pipeline method for transforming requirements specifications in natural language into a model diagram that incorporates Stanford Scene Graph Parsing. The latter is a natural text parser originally used for extracting scene information from image queries. We propose a novel task called common entities and relations extraction that aims to extract all related entities in a text and their relationship, this task will help better model natural text for machine use.  While our entity identification pipeline demonstrates impressive capabilities in handling simple requirement statements, it encounters difficulties when confronted with complex Wikipedia paragraphs. Thus, an improvement of the current work would be to expand this task into all entities extraction including common words. In addition, it would be interesting to expand our evaluation dataset to include more examples from different sources. 









\clearpage
\onecolumn
\begin{landscape}


\end{landscape}
\clearpage
\twocolumn

\bibliographystyle{cas-model2-names}

\bibliography{cas-refs}

\end{document}